\documentclass[conference]{IEEEtran}
\IEEEoverridecommandlockouts

\usepackage{cite}
\usepackage{amsmath,amssymb,amsfonts}
\usepackage{algorithmic}
\usepackage{graphicx}
\usepackage{textcomp}
\usepackage{xcolor}
\def\BibTeX{{\rm B\kern-.05em{\sc i\kern-.025em b}\kern-.08em
    T\kern-.1667em\lower.7ex\hbox{E}\kern-.125emX}}

\usepackage{algorithm}
\usepackage{algorithmic}

\newcommand{\algname}{\mbox{LightDXML}}

\usepackage{newfloat}
\usepackage{listings}
\floatstyle{ruled}
\newfloat{listing}{tb}{lst}{}
\floatname{listing}{Listing}

\usepackage{graphicx}
\usepackage{hyperref}
\usepackage{graphicx}
\usepackage{newfloat}
\usepackage{boldline}
\usepackage{hyperref}
\usepackage{amsmath}
\usepackage{algorithm}
\usepackage{algorithmic}
\usepackage{amssymb}
\usepackage{amsfonts}
\usepackage{bm}
\usepackage{multirow,tabularx,amsbsy}
\usepackage{color}

\begin{document}

\title{Light-weight Deep Extreme Multilabel Classification}

\author{\IEEEauthorblockN{Istasis Mishra\IEEEauthorrefmark{1}, Arpan Dasgupta\IEEEauthorrefmark{1}, Pratik Jawanpuria\IEEEauthorrefmark{2}, Bamdev Mishra\IEEEauthorrefmark{2}, and Pawan Kumar\IEEEauthorrefmark{1}} 
 \IEEEauthorblockA{\IEEEauthorrefmark{1}IIIT Hyderabad, India} 
 \IEEEauthorblockA{\IEEEauthorrefmark{2}Microsoft India
 \\\{istasis.mishra,arpan.dasgupta,pawan.kumar\}@iiit.ac.in and \{pratik.jawanpuria,bamdevm\}@microsoft.com}}

\maketitle

\begin{abstract}

Extreme multi-label (XML) classification refers to the task of supervised multi-label learning that involves a large number of labels. Hence, scalability of the classifier with increasing label dimension is an important consideration. In this paper, we develop a method called {\algname} which modifies the recently developed deep learning based XML framework by using label embeddings instead of feature embedding for negative sampling and iterating cyclically through three major phases: (1) proxy training of label embeddings (2) shortlisting of labels for negative sampling and (3) final classifier training using the negative samples. Consequently, {\algname} also removes the requirement of a re-ranker module, thereby, leading to further savings on time and memory requirements. The proposed method achieves the best of both worlds: while the training time, model size and prediction times are on par or better compared to the tree-based methods, it attains much better prediction accuracy that is on par with the deep learning based methods. Moreover, the proposed approach achieves the best tail-label prediction accuracy over most state-of-the-art XML methods on some of the large datasets\footnote{accepted in IJCNN 2023, partial funding from MAPG grant and IIIT Seed grant at IIIT, Hyderabad, India.}. Code: \url{https://github.com/misterpawan/LightDXML}
\end{abstract}

\section{Introduction}

Extreme multi-label (XML) classification refers to the task of supervised multi-label learning that involves a large number of labels~\cite{agrawal13a,fastxml,sleec_bhatia,PPD_Sparse,jain2019slice}. 
The applications of extreme multi-label classification include a number of short text classification tasks such as matching search engine queries of user to advertiser bid phrases, predicting Wikipedia tags from a document’s title, predicting from a given retail product's name and description, or from frequently bought together products, and showing personalized ads based on the set of web-page titles in a user’s browsing history~\cite{agrawal13a,weston13a,parabel,DiSMEC,xmlcnn,dahiya21a,decaf}.

\paragraph{\textbf{Challenges in extreme multi-label learning}} The extreme classification scenarios~\cite{Bhatia16} have some important properties and patterns that any effective model needs to consider~\cite{agrawal13a,DiSMEC,parabel,deepxml}. Firstly, it has been observed that for most such applications, the average number of positive labels per data point is relatively very small. Specifically, we observe the growth of positive labels per data point to be logarithmic with respect to the number of labels or data points. Secondly, the label frequencies are known to follow Zipf's law \cite{zipf}, which means that only a small fraction of the labels contribute to most of the probability density; these labels are called head labels. On the other hand, the labels that occur in very few data points are termed as tail labels and tail labels are usually large in number. This poses a challenge because training a model for such label means that the model may ``memorize" the dataset for head labels due their high frequency, and the model may not generalize well to predict tail labels. 

\paragraph{\textbf{DeepXML framework for extreme classification}} DeepXML~\cite{dahiya21a} provides a modular deep learning based framework for developing various models for the extreme multi-label classification task. It divides the task broadly into four modules, each of which is trained in a sequential order. The first module aims to solve a proxy task, which involves training feature embeddings on a reduced problem efficiently. The second module takes advantage of the small number of positive labels, and performs a negative sampling of labels to later train against only a small number of ``most confusing" negative labels. The third module seeks to do transfer learning from the intermediate feature embeddings learned in the proxy task in module 1 to the final set of parameters. In the final module, the final set of parameters for the classifier are learned efficiently after doing label shortlisting from the second module. \textbf{Astec} is a method developed in~\cite{dahiya21a} that follows the DeepXML framework. The feature representation is learned based on a clustering of labels, which is then frozen and used as intermediate feature representation during the training of the extreme one-versus-all classifier on the shortlisted labels for each training instance. 

\paragraph{\textbf{Contributions}} 
In this work, we develop a lightweight deep learning based extreme classification approach called {\algname}. 
Compared to DeepXML~\cite{dahiya21a}, the training of intermediate representation in {\algname} is replaced with a method that allows easier and effective retraining, and eliminates the need for freezing the feature representations.

Overall, the proposed approach has two key steps, distinct from the DeepXML framework. First, {\algname} embeds the labels in a low dimensional space, which is later used for negative sampling to find the ``hardest" labels for final classifier training. Secondly, a cyclic training procedure (discussed later in Section \ref{sec:architecture}) is proposed that cycles through shortlisting, classification, and embedding to learn the final model on the negative samples. 

The proposed improvements lead to decrease in train and prediction time for {\algname}, while retaining state-of-the-art generalization accuracy. Moreover, memory required by {\algname}  reduces by two to three folds.  
Our main contributions are summarized below:
\begin{itemize}

    \item Our approach alleviates the requirement for a costly re-ranker, which is typically used in recent deep learning based extreme classification methods.
    \item We illustrate the efficacy of the proposed method {\algname} on several real-world extreme classification datasets~\cite{Bhatia16}. Empirically, our approach improves train and prediction time, and has less memory requirements than  state-of-the-art DeepXML approaches.

\end{itemize}
\section{Related Work}

In recent years, many novel methods have been proposed for the extreme classification problem, and a tremendous amount of progress has been achieved. These methods try to deal with specific problems such as scaling with number of labels, dealing with tail labels or increase the state-of-the-art accuracy. These methods can be loosely classified into following four categories, see \cite{dasgupta2023review} for a detailed review.

\paragraph{\textbf{One-vs-all methods}}
This category includes all the models that split the extreme classification task into multiple independent binary classification problems. State-of-the-art One-vs-all approaches include DiSMEC \cite{DiSMEC}, PPDSparse \cite{PPD_Sparse}, and SLICE \cite{jain2019slice}. These methods are known to be computationally expensive to train, and they have a very high memory requirement. In \cite{parabel}, a method named Parabel is proposed that overcomes this problem by reducing the number of training points in each of the binary classifiers.
\paragraph{\textbf{Tree based methods}}
This category makes use of one or multiple trees built by recursively partitioning the labels and/or features and breaking the original problem down to feasible small scale sub-problems. Examples include SwiftXML \cite{swiftxml}, PfastreXML \cite{pfastrexml}, Bonsai \cite{bonsai}, FastXML \cite{fastxml}, and CRAFTML \cite{craftml_siblini}. Although, these methods have fast training and prediction time, they suffer in prediction accuracy.
\paragraph{\textbf{Embedding based methods}}
These methods rely on low-dimensional embeddings to approximate the label space. Examples include AnnexML \cite{annexml} and SLEEC \cite{sleec_bhatia}. These methods, like the tree based methods, have fast training and prediction times, but have low prediction accuracy \cite{naram2022}.
\paragraph{\textbf{Deep learning based methods}}
This category includes models that rely on deep learning methods to capture complex non-linear correlations between the features and output labels. These methods, at times, also rely on the usage of binary classifiers like One-vs-all methods and embeddings, for example, one of the first methods to do so was XML-CNN \cite{xmlcnn}; it relies on the use of CNNs on word embeddings to calculate the low-dimensional embeddings for features. A new method APLC-XLNet \cite{aplc-xml} is another deep learning approach that performs very well on smaller datasets, but performs poorly on the bigger datasets. The recently proposed DeepXML framework provides a framework for formulating a deep learning based method to solve the XML problem. Astec is the first implementation and it gives a noticeable improvement over the state-of-the-art precision scores. Some of the later implementations include DECAF \cite{decaf} and ECLARE \cite{eclare}. All the methods in this category give better accuracy than the other methods, but have a very high training time, memory requirement, and prediction time. 
Our method \algname ~is an instance of this category.

\subsection*{ASTEC: An Implementation in DeepXML Framework}
In this section, we briefly describe the Astec implementation of DeepXML framework. As mentioned before, {\algname} is based on a modification of the DeepXML framework~\cite{dahiya21a}. Therefore, we also compare {\algname} primarily with the architecture of Astec, the algorithm derived using the DeepXML framework in~\cite{dahiya21a}.
Let $L$ be the number of labels, $V$ be the vocabulary size of the feature vectors, $N$ be the number of data points in the training set, and $N_t$ be the number of data points in the testing set. The notation $[x, y]$ is used to denote all integers between $x$ and $y$ inclusive. Each data point is represented as a pair of features and labels $(\bm{x_i}, \bm{y_i})$, where $\bm{x_i} \in \mathbb{R}^V$ is a sparse TF-IDF weighted bag-of-words representation of the data point and $\bm{y_i} \in [0, 1]^L$ is the ground truth label vector with $y_{il} = 1$ for all $l \in [1, L]$ if $l$ is a relevant positive label for the data point $i$ and $y_{il} = 0$, otherwise. Astec has the following components:

\subsubsection{Module 1}
A feature encoder $\bm{\mathcal{Z}}$ that maps a data point $\bm{x_i}$ onto a dense $D$ dimensional representation $\hat{\bm{x_i}}$, i.e., $\bm{\mathcal{Z}} : \bm{x_i} \longrightarrow \hat{\bm{x_i}} \in \mathbb{R}^D$.
A surrogate objective is used to train intermediate feature representations.  Astec adopts label clustering for its surrogate task. Label centroids, defined as $\boldsymbol{\mu}_{l}^{s}=\frac{\hat{\boldsymbol{\mu}}_{l}^{s}}{\left\|\hat{\boldsymbol{\mu}}_{l}^{s}\right\|_{2}}$, are used to cluster the labels, where, $\hat{\boldsymbol{\mu}}_{l}^{s}=\frac{1}{\left|\mathcal{P}_{l}\right|} \sum_{i \in \mathcal{P}_{l}} \mathbf{x}_{i}$, and $\mathcal{P}_{l}:=\left\{i: y_{i l}=+1\right\}$ is the set of documents for which label $l$ is relevant. The $\hat{L}$ number of clusters are obtained using 2-means++ algorithm. These clusters are treated as meta-labels, and new (meta) label vectors $\hat{\mathbf{y}}_{i} \in\{-1,+1\}^{\hat{L}}$ are created for each training document as $\hat{y}_{i k}=+1$ for documents $i$ tagged with at least one label in cluster $k$ and $\hat{y}_{i k}=-1$ otherwise. Predicting the relevant clusters for a given document is then taken as the surrogate task for Module 1. This is fundamentally different from \algname. \algname \ does not require a surrogate task, as the goal is to learn label embeddings instead of intermediate feature embeddings.

\subsubsection{Module 2}
In Module 2, Astec samples the $\mathcal{O}(\log L) \leq 500$ hardest negative labels using two Approximate Nearest Neighbor Structures (ANNS) \cite{malkov} along with 50 randomly selected negative labels. 
Astec freezes the intermediate representations after completion of Module 1. 
\noindent However, this step is different in LightDXML. In LightDXML, we adopt a strategy of re-sampling for the negative sample after training the extreme classifier (see Module 3 below)  for a fixed number of epochs to account for changing feature representations due to final embeddings and classifier training.

\subsubsection{Module 3}
In this module, Astec uses transfer learning to prepare a final model called extreme classifier for training using the intermediate feature representations obtained in Module 1 and using shortlisted labels in Module 2. More specifically, the transfer to final feature representations from intermediate feature representations is done using a composition of RELU and a linear transform matrix ${\bf R}$ in Astec.

\subsubsection{Module 4}
Finally, Astec trains the parameters $\bm{W}$ of a extreme classifier model to make final predictions in this module using the shortlist from Module 2 and transferred features in Module 3. This step is same in LightDXML.

\subsubsection{Re-ranker}
Astec improves its accuracy by learning a novel re-ranker as follows. Astec's predictions $\hat{y}_{i} \in\{-1,+1\}^{L}$ are obtained for each training point $i.$ The negative labels wrongly classified to be in the top-$k$ are included in a list of negative samples.

A re-ranker is then trained using the negative samples along with the positive labels for each point to eliminate mis-predictions. The re-ranker has an architecture similar to the feature encoder $\bm{\mathcal{Z}},$ but with own separate learnable parameters.
LightDXML achieves a similar level of accuracy without the re-ranker, thus, cutting down on training time, prediction time, and model size.

\section{{\algname}: Architecture}
\label{sec:architecture}

\begin{figure}[t]
    \centering
    \includegraphics[scale=0.42]{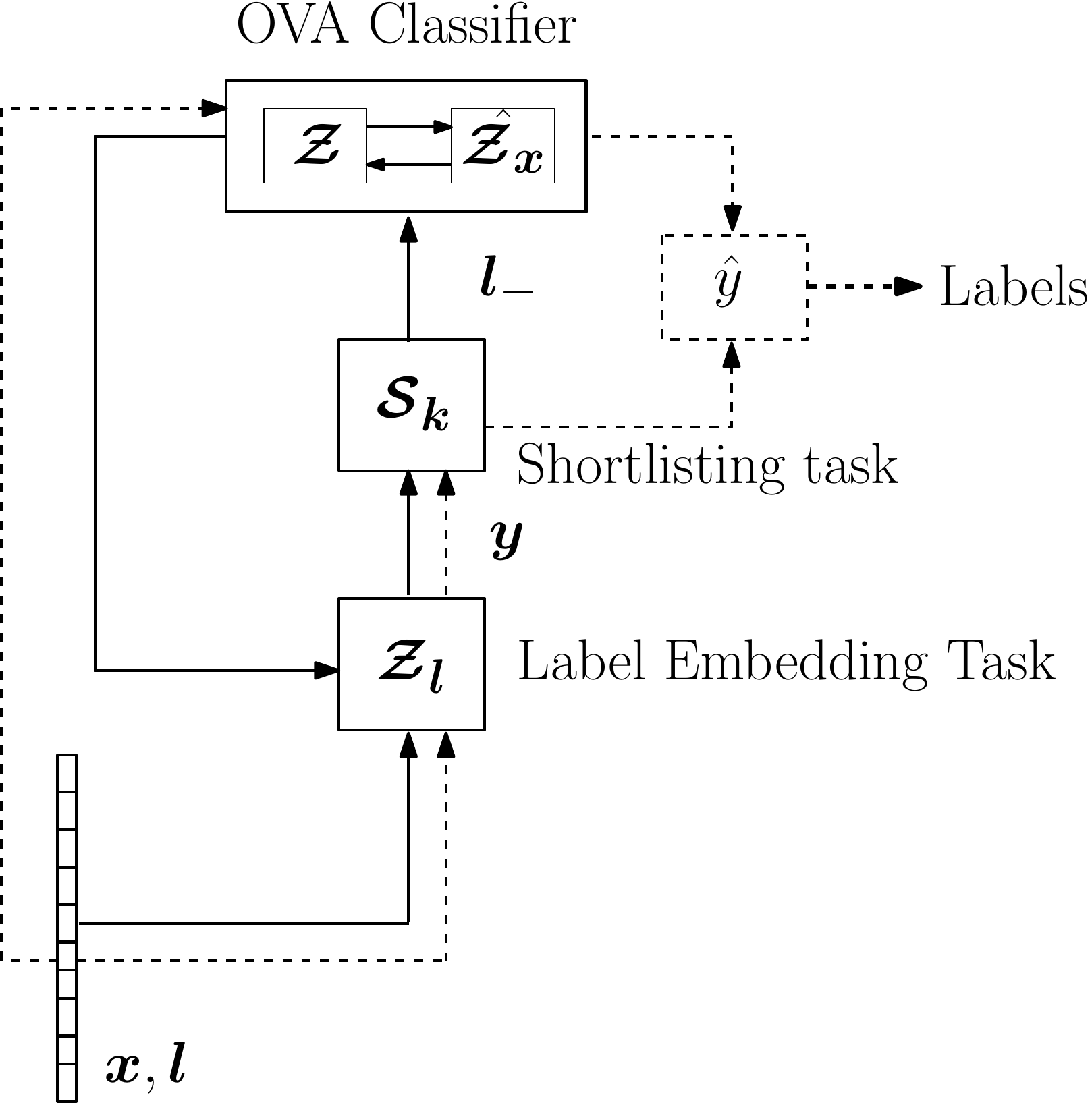}
    \caption{\label{fig:dss} The architecture of {\algname}: Please refer to the text in Section \ref{sec:architecture} for details. The block $\bm{\mathcal{Z}_l}$ is the label encoding architecture that generates dense label embeddings. The block $\bm{\mathcal{S}_k}$ additionally generates the $k$ negative labels. Finally, the block $(\bm{\mathcal{Z}}, \hat{\bm{\mathcal{Z}_x}})$ is the extreme classifier that also generates a new set of data feature embeddings. $\hat{\bm{\mathcal{Z}_x}}$ is then fed into the block ${\bm{\mathcal{Z}_l}}$ for alternating optimization over the blocks. This is continued for a certain number of iterations. Here, the solid lines shows the training phase and the dotted lines show the prediction phase. The OVA stands for one-versus-all.}
\end{figure}

This section introduces relevant notations and describes {\algname}'s overall architecture. Figure \ref{fig:dss} shows the overall architecture of {\algname}. The various steps are in mentioned in Algorithm \ref{alg:propose}.

Given a loss function $\ell$, a standard strategy is to jointly learn $\bm{\mathcal{Z}_x}$ (feature embedding of data points) and $\bm{\mathcal{Z}}$ (classifier), i.e.,Given a loss function $\ell$, a standard strategy is to jointly learn $\bm{\mathcal{Z}_x}$ (feature embedding of data points) and $\bm{\mathcal{Z}}$ (classifier), i.e.,

\begin{equation}
\label{eq:full_problem}
    \underset{\bm{\mathcal{Z}_x}, \bm{\mathcal{Z}}}{\text{argmin}}~ \mathcal{L}(\bm{\mathcal{Z}_x}, \bm{\mathcal{Z}}) = \underset{\bm{\mathcal{Z}_x}, \bm{\mathcal{Z}}}{\text{argmin}}~\dfrac{1}{N} \sum\limits_{i=1}^N \sum\limits_{j=1}^L \ell(\bm{x_i}, y_{ij}; \bm{\mathcal{Z}_x}, \bm{\mathcal{Z}}).
\end{equation}
Problem (\ref{eq:full_problem}) requires dealing with all the labels, and this makes the problem not viable even for moderate size datasets. Instead of directly solving the problem (\ref{eq:full_problem}), the DeepXML framework~\cite{dahiya21a} proposes to find a smaller set of labels by solving a set of proxy tasks, and then using the smaller set of labels to learn the classifier. 

The {\algname} method also follows the DeepXML framework, but we do {\it three} modifications: 1) learning of label embeddings directly from feature vectors instead of feature embeddings (unlike in Astec) 2) not freezing of data feature representations 3) iterating over label embeddings, shortlisting of labels, and training the extreme classifier. 
To this end, the key steps of {\algname} are explained in detail below. 
{\tiny
\begin{algorithm}[tb]
   \caption{Various steps in {\algname}.}
   \label{alg:propose}
   \small
\begin{description}
    \item [Step 1] Every data point $\bm{x_i}$ is mapped to the metric space $\mathcal{M} = \mathbb{R}^D$ denoted by $\hat{\bm{x}}_i$, using a TF-IDF weighted combination of FastText embeddings.
    
    \item [Step 2] A label encoder $\bm{\mathcal{Z}_l}$ maps each label to the same metric space $\mathcal{M}$, i.e., $\bm{\mathcal{Z}_l}: \bm{l_j} \to \hat{\bm{l_j}} \in \mathcal{M},$ where $\bm{l_j}$ is the label centroid of the $j^{th}$ label. We use the Euclidean distance metric in the space $\mathcal{M}$. 
    
    \item [Step 3] Using the label encoding $\bm{\mathcal{Z}_l}$, a shortlist $\bm{\mathcal{S}_k}(\bm{x_i})$ is created to search for the top $k$ labels using a sub-linear search structure, i.e, $\bm{\mathcal{S}_k}: \bm{x_i} \to [\bm{l_{i_1}},...,\bm{l_{i_k}}] \in [1, L]^k$.  
    
    \item [Step 4] A sparse one-vs-all extreme classifier $\bm{\mathcal{Z}}:  \bm{x_i} \to \mathbb{R}^L$ is trained to predict the final labels. This step modifies the feature representation of data points to generate $\hat{\bm{\mathcal{Z}_x}}$.
    
    \item [Step 5] Iterate over steps 2, 3, and 4 until (validation) accuracy is maximized. This completes the training process.
    
    \item [Step 6] The final predictions are made as an ensemble of the shortlist and the extreme classifier.

\end{description} 
\end{algorithm}
}
\normalsize

In Algorithm \ref{alg:propose}, the core idea of {\algname} in {\bf Step 1} is to obtain feature representations for each data point $\bm{x_i} \in \mathbb{R}^D$ using TF-IDF weighted combination of FastText embeddings.
Unlike in Astec, in {\bf Step 2}, we additionally learn label embeddings as well. 
Specifically, the goal is now to train label encoder $\bm{\mathcal{Z}_l}$ to minimize the mean squared error between feature representations $\bm{x}_i$ (obtained from {\bf Step 1}) and label embedding $\hat{\bm{l}}_j$ if the $j^{th}$ label is positive for the data point $i$. By letting $\bm{\mathcal{P}_i}$ be the set of all positive labels for the $i^{th}$ data point, i.e., $\bm{\mathcal{P}} = \{j: y_{ij}=1\}$ the objective $\mathcal{L}_l$ effectively becomes
\begin{equation}\label{eq:label_training}
\mathcal{L}_l(\bm{\mathcal{Z}_l})= \dfrac{1}{N} \sum_{i=1}^{N} \sum_{\substack{j=1\\y_{ij}=1}}^L \log(1 + ||\bm{x_i} - \bm{\mathcal{Z}_l}(j)||^2_2).
\end{equation}

In {\bf Step 3}, {\algname} finds the top $k = \bm{\mathcal{O}}(\log(L))$ most hard negative labels for each data point using the shortlist $\bm{\mathcal{S}_k}$ by finding the $k$ nearest label points $\hat{\bm{y_j}}$ to the feature representation $\bm{x_i}$. Like in Astec, this is achieved in $\bm{\mathcal{O}}(\log(L))$ time per data point using an ANN data structure \cite{malkov}. Using $\bm{\mathcal{S}_k}$, we now have access to both positive and negative labels for a data point.

In {\bf Step 4}, making use of $\bm{\mathcal{S}_k}$, {\algname} learns the one-vs-all classifier $\bm{\mathcal{Z}}$ by jointly learning data feature embeddings $\hat{\bm{\mathcal{Z}_x}}$, and training on the union of set of positive labels and a set of shortlisted negative labels (instead of training on all labels), i.e.,
\begin{equation}\label{eq:small_problem}
\begin{array}{lll}
\underset{\hat{\bm{\mathcal{Z}_x}}, \bm{\mathcal{Z}}}{\operatorname{argmin\;}} \hat{\mathcal{L}}(\hat{\bm{\mathcal{Z}_x}}, \bm{\mathcal{Z}})  =\sum\limits_{i=1}^N \: \sum\limits_{j \in \bm{\mathcal{P}_i} \cup \bm{\mathcal{S}_k}(\bm{x_i})} \ell(\bm{x_i}, y_{ij}; \: \hat{\bm{\mathcal{Z}_x}}, \bm{\mathcal{Z}}).
\end{array}
\end{equation}
It should be noted that Problem (\ref{eq:small_problem}) is more efficient to solve than the original problem (\ref{eq:full_problem}), due to training on a much reduced set of labels. We note that optimizing $\hat{\bm{\mathcal{L}}}$ in (\ref{eq:small_problem}) changes the feature representations $\hat{\bm{\mathcal{Z}_x}},$ which are now possibly very different from the initial features ${\bm{x}}$ which in turn changes label embeddings $\bm{\mathcal{Z}_l}.$ Consequently, the shortlisting of labels $\bm{\mathcal{S}_k}$ in {\bf Step 3} is no longer valid because it should change with updated ${\bm{\mathcal{Z}_l}}.$

Astec solves this problem by ensuring that the intermediate feature representations obtained from the proxy task are close to the final features (by restricting the spectral norm during feature transfer). The intermediate representations are then frozen while optimizing $\hat{\bm{\mathcal{L}}}$. This creates a loss in performance, which Astec regains by training a re-ranker after the classification task. However, it further increases the training time and model size.
On the other hand, {\algname} does not freeze the data feature representations, rather it iteratively optimizes $\hat{\bm{\mathcal{L}}}$ in \eqref{eq:small_problem} and $\bm{\mathcal{L}_l}$ in \eqref{eq:label_training} in a loop. This is our {\bf Step 5}. As shown in experiments later, this preserves the accuracy of the proposed model, while decreasing the training time (as it eliminates the need for a re-ranker) and reduces model size to half or even a third of that for Astec. {\bf Step 6} is our prediction step.

\section{{\algname}: Details}

In this section, we discuss the details of the implementation of {\algname}. The time complexity for each step is shown in Table \ref{tab:time_complexity}.

\subsection{Obtaining Features from Sample Text} 

The features are obtained from the text by using a TF-IDF weighted bag-of-words method. The features for each word are obtained from the FastText embeddings. This step is fundamentally different from that performed in Astec. Astec trains a proxy feature encoder 
by initializing with these feature vectors, which is further used for label shortlisting.
On the other hand, \algname \ does not need a feature encoder, because the label embeddings trained later in {\bf Step 2} are used for label shortlisting instead.

\subsection{Training Label Embeddings} 

{\algname} makes use of label embeddings, which are learned efficiently. The labels are embedded in the same $D$ dimensional metric space $\mathcal{M} = \mathbb{R}^D$ as the features. The label encoder $\bm{\mathcal{Z}_l}$ learns $D$ dimensional dense label embeddings for each label $\hat{\bm{l_j}} \in \mathbb{R}^D$ for all $j \in [1, L]$ in order to minimize the loss \eqref{eq:label_training}. The final embeddings are of the form 
\begin{equation}
\begin{array}{lll}
\hat{\bm{y_j}} &= \bm{\mathcal{Z}_l}(j) \\
&= \text{ReLU}(\bm{W_l^{(2)}}(\bm{W_l^{(1)}\mu_j} + \bm{b_l^{(1)}}) + \bm{b_l^{(2)}})\\ 
&= \text{ReLU}((\bm{W_l^{(2)}}\bm{W_l^{(1)})\mu_j} + (\bm{W_l^{(2)}}\bm{b_l^{(1)}} + \bm{b_l^{(2)}}))\\ 
&= \text{ReLU}(\bm{W_l\mu_j} + \bm{b_l}),
\end{array}
\end{equation}
where
\begin{equation}
\begin{array}{lll}
\bm{\mu_j}&=\dfrac{1}{|\bm{\mathcal{P}_j}|}\sum_{i \in \bm{\mathcal{P}_j}} \bm{x_i},\quad 
\bm{\mathcal{P}_j} &= \{i : l_{ij}=1\},
\end{array}
\end{equation}
where $$\bm{W_l} = \bm{W_l^{(2)}} \bm{W_l^{(1)}} \in \mathbb{R}^{D \times D}$$ are the trainable weights and $$\bm{b_l} = \bm{W_l^{(2)}b_l^{(1)} + b_l^{(2)}}  \in \mathbb{R}^D$$ are the biases. Thus, the label encoder $\bm{\mathcal{Z}_l}$ is a sequence of two fully connected layers, i.e, $\bm{W_l^{(1)}} \in \mathbb{R}^{\hat{D} \times D}$, $\bm{W_l^{(2)}} \in \mathbb{R}^{D \times \hat{D}}$, $\bm{b_l^{(1)}} \in \mathbb{R}^{\hat{D}}$ and $\bm{b_l^{(2)}} \in \mathbb{R}^D,$ here $\hat{D}$ is chosen to be $\mathcal{O}(D)$ to make {\algname} efficient.
{\algname} trains $\bm{\mathcal{Z}_l}$ as the label embedding task 
instead of the indirect method of obtaining label embeddings from the feature embedding task as mentioned before. This step is followed by shortlisting the ``most confusing" negative labels while learning the extreme classifier $\bm{\mathcal{Z}}$.

{\tiny
\begin{table}[t]
\caption{\label{tab:time_complexity}Comparison of time complexities between Astec and {\algname}. $\Bar{L}$ represents the number of positive labels per sample ($\Bar{L} \ll L$) and $\hat{L}$ represents the number of meta labels in the Astec surrogate task. In general, $\bar{L}$ is lesser than $\hat{L}$ considering the experimental values. It should be noted that Step 1 of \algname \ corresponds to Module 1 of Astec, Step 3 corresponds to Module 2, and Step 4 corresponds to Module 3 and 4 combined.}
   \setlength{\tabcolsep}{0.5em}
    \centering
    \small
    \begin{tabular}{|c|c|c|}
        \cline{1-3}
        \textbf{Steps}  &\textbf{Astec}   &\textbf{{\algname}}\\
        \cline{1-3}
        1   	&$\mathcal{O}(L \log(\hat{L}) + ND\hat{L})$ &$\mathcal{O}(ND)$\\
        2 	&Not used 	 &$\mathcal{O}(ND\Bar{L})$\\
        3 	&$\mathcal{O}(ND\log(L))$ 	 &$\mathcal{O}(ND\log(L))$\\
        4 	&$\mathcal{O}(ND\log(L))$ 	&$\mathcal{O}(ND\log(L))$\\
        Re-ranker &$\mathcal{O}(ND\log(L))$ 	&Not needed\\
        \cline{1-3}
    \end{tabular}
        
    \end{table}
}
 \normalsize

\subsection{Shortlisting} 
The shortlist $\bm{\mathcal{S}_k}$ is responsible for shortlisting the top $k$ most confusing negative labels for each data point $\bm{\mathcal{S}_k} = [\bm{\hat{y}_{i_1}},...,\bm{\hat{y}_{i_k}}]$ based on label embeddings $\bm{\hat{l}_i}$. 
Since the feature representations and label embeddings are trained on the same metric space $\mathcal{M}$, we use cosine similarity between the feature representations and label embeddings to find top $k$ labels. In particular, the shortlist is found as follows:
\begin{equation}\label{eq:shortlist}
\bm{\mathcal{S}_k}(\hat{\bm{x_i}})_t =
\begin{cases}
\underset{j \in [1, L]}{\operatorname{argmin\;}} 1 - \dfrac{\bm{\hat{\bm{x_i}}}^T\bm{\hat{\bm{y_j}}}}{||\bm{\hat{\bm{x_i}}}||_2||\bm{\hat{\bm{y_j}}}||_2}, \text{if } t=1 \\
\underset{j \in [1, L]}{\operatorname{argmin\;}} \left\{1 - \dfrac{\bm{\hat{\bm{x_i}}}^T\bm{\hat{\bm{y_j}}}}{||\bm{\hat{\bm{x_i}}}||_2||\bm{\hat{\bm{y_j}}}||_2} :\right. \\ \left. \dfrac{\bm{\hat{\bm{x_i}}}^T\bm{\hat{\bm{y_j}}}}{||\bm{\hat{\bm{x_i}}}||_2||\bm{\hat{\bm{y_j}}}||_2} \geq \right.  \left.
\dfrac{\bm{\hat{\bm{x_i}}}^T\bm{\mathcal{S}_k}(\hat{\bm{x_i}})_{t-1}}{||\bm{\hat{\bm{x_i}}}||_2||\bm{\mathcal{S}_k}(\hat{\bm{x_i}})_{t-1}||_2}\right\}, \\ \text{otherwise}.
\end{cases}
\end{equation}

Here, $\bm{\mathcal{S}_k}(\hat{\bm{x_i}})_t$ refers to the $t$-th shortlisted label for point $\bm{\hat{x_i}}$.
The shortlisting (\ref{eq:shortlist}) is done efficiently in $\mathcal{O}(\log(L))$ time using an ANNS data structure like Hierarchical Navigable Small World (HNSW) \cite{malkov}.
Later, during prediction, this label shortlist $\bm{\mathcal{S}_k}$, along with the label encoder $\bm{W_l}$ can also be used as a classifier, where the shortlisted labels are taken as predicted labels and the $j_{th}$ label is given the score of 
\[
(\hat{y}_{ij})_{\text{hnsw}} =
\begin{cases} \dfrac{\bm{\hat{\bm{x_i}}}^T\hat{\bm{y_j}}}{||\bm{\hat{\bm{x_i}}}||_2||\hat{\bm{y_j}}||_2}, &\hat{\bm{y_j}} \in \bm{\mathcal{S}_k}(\hat{\bm{x_i}})\\
0, &\text{otherwise}.
\end{cases}
\]
Since, $k = \mathcal{O}(\log(L))$, the predicted label matrix $\hat{\bm{l}}$ is very sparse.

\subsection{Extreme Classifier} 

The extreme classifier in {\algname} is a one-vs-all classifier that learns $D$ dimensional data for each label. Thus, the classifier $\bm{\mathcal{Z}}$ is parameterized by $\bm{W} = [\bm{w_1},...,\bm{w_L}] \in \mathbb{R} ^ {D \times L}$. The loss used is \eqref{eq:small_problem} uses the label shortlist, thus, does not require iterating over the entire label set.

\subsection{Training}

As mentioned before, {\algname} deviates from the DeepXML framework in the learning of label encoder $\bm{\mathcal{Z}_l}$, final classifier $\bm{\mathcal{Z}}$, and final feature embeddings $\hat{\bm{\mathcal{Z}_x}}$. In particular, these three parameters are trained in an cyclic fashion. 
As mentioned before, in Module 3 of Astec, due to transfer learning of intermediate features to final feature embeddings, the negative samples obtained before may become invalid. The cyclic procedure proposed in \algname \ attempts to solve this problem; it does not increase the training time much as observed in our experiments. 
Overall, $\bm{\mathcal{Z}}$ and $\hat{\bm{\mathcal{Z}_x}}$ are trained jointly for $e_{\text{model}}$ epochs, and $\bm{\mathcal{Z}_l}$ is trained for $e_{\text{label}}$ epochs every $\hat{e}_{\text{label}}$ and once before the first epoch. Hence, the total number of epochs that the model runs for is $$e_{\text{model}} + e_{\text{label}}(\left\lfloor{\dfrac{e_{\text{model}}}{\hat{e}_{\text{label}}}}\right\rfloor+1)$$.
First, $\hat{\bm{\mathcal{Z}_x}}$ is initialized using either random initialization or FastText embeddings. After that, the trainable parameters in $\bm{\mathcal{Z}_l}$ as described before are trained using the loss (\ref{eq:label_training}).

After this training, the shortlist $\bm{\mathcal{S}_k}$ of labels is calculated for each data point, where $k$ is a hyperparameter that is typically kept between 400 and 700.
Finally, $\bm{\mathcal{Z}}$ and $\bm{\hat{\mathcal{Z}}_x}$ are trained jointly using the following binary cross entropy loss with logits
\begin{align}
%\begin{array}{lll}
\hat{\bm{\mathcal{L}}}(\hat{\bm{\mathcal{Z}_x}}, \bm{\mathcal{Z}}) 
  &= \sum_{i=1}^N \sum_{j \in \bm{\mathcal{P}_i} \cup \bm{\mathcal{S}_k}(\bm{x_i})}-y_{ij}\log(\sigma((\hat{y_{ij}})_{\text{clf}}))) \\
 &- (1-y_{ij})(\log(1-\sigma((\hat{y}_{ij})_{\text{clf}}))),
%\end{array}
\end{align}
where $(\hat{y}_{ij})_{\text{clf}}$ is the predicted score of the classifier and $\sigma(x)=\frac{1}{1+e^{-x}},$ which is the well-known sigmoid activation function.

\subsection{Prediction}\label{sec:prediction}

The label embeddings are used to pick top $k$ number of labels and prediction is done using the classifier only on the labels in the shortlist. Hence, prediction also takes ${\mathcal{O}}(D\log L)$ time. The labels are given a final score of
\begin{align*}
\hat{y}_{ij} = \beta \sigma((\hat{y}_{ij})_{\text{clf}}) + (1-\beta)(\sigma((\hat{y}_{ij})_{\text{hnsw}})),
\end{align*}
where $\beta$ is a hyperparameter that weighs the prediction from the shortlist with that of the extreme classifier.

{\tiny
\begin{table*}
\begin{center}
%\small 
\caption{\label{tab_results}Test times are reported in milliseconds. ``Embd" stands for embedding based models, ``Tree" are tree based models and ``DL" stands for deep learning models. AnnexML \cite{annexml} is implemented in C, PfastreXML \cite{pfastrexml} and fastXML \cite{fastxml} in C++, SLEEC \cite{sleec_bhatia} in MATLAB, and Astec and {\algname} in Python.  The highest value is in \textbf{bold}. The values for which {\algname} achieves better than all the known state-of-the-art methods mentioned also in \cite{Bhatia16} for P@K or PSP@K values are depicted with a \framebox{\textbf{bounding box}}. For Astec, we consider results in table. Here NA stands for Not Available. We tried to choose the best hyper-parameters for the methods compared. Most comparisons are made on the same machine ensuring same hardware conditions. However, for larger problems, Parabel* and Bonsai* data are taken from \cite{Bhatia16}. Bold and underline denotes best train times for methods run our machine. Note that both Parabel and Bonsai have poor P and PSP scores. LightXML \cite{jiang2021lighxml} and APLC-XLNet \cite{aplc-xml} achieve high $P$ scores for the first three datasets from top, but they don't report train time and their model sizes are very large, and hence are excluded from comparisons.}%

\small
\begin{tabular}{|c|c|ccc|ccc|ccc|}
\hline 
\multirow{2}{*}{\textbf{Class}}
 &\multirow{2}{*}{\textbf{Method}}
 &\multicolumn{3}{c|}{\textbf{Precision Scores}}
 &\multicolumn{3}{c|}{\textbf{Propensity Scores}}
 &\multicolumn{3}{cV{2.7}}{\textbf{Performance }}\\
 
 \cline{3-11}
    & &\textbf{P@1}    &\textbf{P@3}    &\textbf{P@5}     &\textbf{PSP@1}    &\textbf{PSP@3}    &\textbf{PSP@5}    &\textbf{Train}  &\textbf{Size}    &\textbf{Test}\\

\hline

\multicolumn{11}{V{2.7}cV{2.7}}{\textbf{EURLex-4K}}\\

\hline

\multirow{2}{*}{\textbf{Emb}} &AnnexML &80.04 &64.26 &52.17 &34.25 &39.83 &42.76 &146s &85.8M & NA\\
&SLEEC &79.36 &64.68 &52.94 &24.10 &27.20 &29.09 &715s &120M &0.86\\

\hline

\multirow{2}{*}{\textbf{Tree}} &PfastreXML &70.07 &59.13 &50.37 &26.62 &34.16 &38.96 &347s &255M &\textbf{0.13}\\

&FastXML &70.94 &59.90 &50.31 &33.17 &39.68 &41.99 &\textbf{18s} &218M &0.47\\

&Bonsai &81.22 &67.80 &55.38 &36.91 &44.20 &46.63 &84s &23.5M &2.07\\

\hline

\multirow{2}{*}{\textbf{DL}} &Astec &80.81 &67.50 &55.66 &37.75 &43.80 &47.01 &136s &52.5M &0.95\\

&{\algname} &\textbf{82.38} &\textbf{67.94} &\textbf{56.18} &\textbf{38.51} &\textbf{44.57} &\textbf{47.39} &75s &\textbf{23.2M} &0.42\\

\hlineB{2.7}

\multicolumn{11}{V{2.7}cV{2.7}}{\textbf{AmazonCat-13K}}\\

\hlineB{2.7}

\multirow{2}{*}{\textbf{Emb}} &AnnexML &93.54 &78.37 &63.31 &49.04 &61.13 &69.64 &4289s &18.4G & 0.46\\
&SLEEC &89.94 &75.82 &61.48 &46.75 &58.46 &65.96 &6937s &5.9G &0.64\\
\hline
\multirow{3}{*}{\textbf{Tree}} &PfastreXML &85.60 &75.23 &62.87 &\textbf{69.52} &\textbf{73.22} &\textbf{75.48} &1719s &18.9G & 0.50\\
&FastXML &93.09 &78.18 &63.58 &48.31 &60.26 &69.30 &1696s &18.3G &0.44\\
&Bonsai* &92.98 &79.13 &64.46 &51.30 &64.60 &72.48 & {\bf 75.6s} &0.55G &NA\\
\hline
\multirow{2}{*}{\textbf{DL}} &Astec &87.54 &74.87 &60.88 & 58.02 & 64.20 & 70.32 &7807s &2.3G &1.00\\
&{\algname} &{\textbf{94.76}} &{\textbf{79.06}} &{\textbf{64.96}} &59.88 &67.40 &71.43 &\underline{\textbf{1614s}} &\textbf{291.4M} &\textbf{0.26}\\
\hlineB{2.7}

\multicolumn{11}{V{2.7}cV{2.7}}{\textbf{Wiki10-31K}}\\

\hlineB{2.7}
\multirow{2}{*}{\textbf{Emb}} &AnnexML &86.22 &73.28 &64.19 &11.90 &12.76 &13.58 &1404s &634.9M & NA\\
&SLEEC &85.88 &72.98 &62.70 &11.14 &11.86 &12.40 &756s &1157.1M &0.66\\
\hline

\multirow{2}{*}{\textbf{Tree}} &PfastreXML &83.57 &68.61 & 59.10 &19.02 &18.34 &18.43 &66s &482M &2.06\\
&FastXML &83.03 &67.47 &57.76 &9.80 &10.17 &10.54 &\textbf{59s} &482M &0.69\\
&Bonsai &84.61 &73.13 &64.36 &11.85 &13.47 &14.69 &3077s &408.3M &1.14\\
\hline

\multirow{2}{*}{\textbf{DL}} &Astec &80.79 &50.51 &37.13 &9.98 &7.28 &6.16 &237s &383.9M &1.29\\
&{\algname} &\textbf{86.31} &\textbf{73.45} &\textbf{64.37} &\framebox[1.1\width]{\textbf{23.43}} &\framebox[1.1\width]{\textbf{21.68}} &\framebox[1.1\width]{\textbf{21.18}} &168s &\textbf{249.1M} &\textbf{0.57}\\
\hlineB{2.7}

\multicolumn{11}{V{2.7}cV{2.7}}{\textbf{WikiSeeAlsoTitles-350K}}\\

\hlineB{2.7}
\multirow{1}{*}{\textbf{Emb}} &AnnexML &14.96	&10.20	&8.11 &5.63 &7.04 &8.59 &1800s &3.5G & NA\\
\hline
\multirow{4}{*}{\textbf{Tree}} &PfastreXML &15.09	&10.49	&8.24 &9.03 &9.69 &10.64 &1800s &5.2G &8.23\\
&FastXML &13.36 &8.68 &6.55 &5.25 &6.05 &6.75 &2157s &4.07G &16.08\\
&Bonsai* &17.95 &12.27 &9.56 &8.16 &9.68 &11.07 &1656s & {\bf 0.25G} & NA\\
&Parabel* &17.24 &11.61 &8.92 &7.56 &8.83 &9.96 & {\bf 216s} & {\bf 0.43G} & NA\\
\hline
\multirow{2}{*}{\textbf{DL}} &Astec &19.76 &13.58 &10.52 &9.91 &\framebox[1.1\width]{\textbf{12.16}} &\framebox[1.1\width]{\textbf{14.04}} &5155s &2.5G &0.61\\
&{\algname} &\framebox[1.1\width]{\textbf{19.92}} &\framebox[1.1\width]{\textbf{13.69}} &\framebox[1.1\width]{\textbf{11.02}} &\framebox[1.1\width]{\textbf{10.35}} &{11.30} &{12.55} &\underline{\textbf{1430s}} &\underline{\bf 1.5G} &\textbf{0.36}\\
\hlineB{2.7}
\multicolumn{11}{V{2.7}cV{2.7}}{\textbf{AmazonTitles-670K}}\\
\hlineB{2.7}
\multirow{1}{*}{\textbf{Emb}} &AnnexML &35.31 &30.90 &27.83 &17.94 &20.69 &23.30 &3612s &2.99G & NA\\
\hline
\multirow{4}{*}{\textbf{Tree}} &PfastreXML &32.88 &30.54 &28.80 &26.61 &27.79 &29.22 &3996s &5.32G &25.11\\
&FastXML &33.74 &30.56 &28.19 &18.24 &21.77 &25.04 &3780s &5.05G &0.78\\
&Bonsai* &38.46 & 33.91 & 30.53 & 23.62 & 26.19 & 28.41 & {\bf 1908s} & {\bf 0.66G} & NA\\
&Parabel* &38.00 & 33.54 & 30.10 & 23.10 & 25.57 & 27.61 & {\bf 324s} & {\bf 1.06G} & NA\\
\hline
\multirow{2}{*}{\textbf{DL}} &Astec &\framebox[1.1\width]{\textbf{40.07}} &35.61 &32.65 &25.03 &28.17 &\textbf{31.01} &4267s &3.64G &2.24\\
&{\algname} &39.12 &\textbf{35.88} &\textbf{33.96} &{\textbf{27.23}} &{\textbf{28.83}} &30.56 & \underline{\textbf{3535s}} &\underline{\textbf{2.84G}} &\textbf{0.66}\\
\hlineB{2.7}
\end{tabular}

%\egroup
\end{center} 
\end{table*}
}

\normalsize

\subsection{Remark on the Role of Re-ranker}

Re-ranker is an extra supplementary module learned in addition to the original classifier to improve the accuracy of the model. It can be used with several possible goals in mind such as removing misclassification due to some inherent assumption in the model. For example, PfastreXML \cite{jain2016extreme} used re-ranker for enhancing the performance. Without a re-ranker, in \cite{jain2016extreme} a lower probability of prediction of tail labels by the internal tree nodes was found. The usage of re-ranker solves this as they are tuned specifically for predicting tail labels with higher accuracy. DEFRAG \cite{jalan2019accelerating} also suggests a re-ranker algorithm called REFRAG based on feature agglomeration, which can enhance the performance on tail labels. 
In Astec the ANN search used for negative sampling is not trained on the final features but on the surrogate task. This loss in performance is countered by the usage of a re-ranker, which considers the wrongly predicted negative labels as a part of the negative sample. This improves the accuracy for Astec, but we found it to be costly to perform. On the other hand, LightDXML trains the final model on the label embeddings which are valid for the final feature representation thanks to the cyclic update procedure followed in Algorithm \ref{alg:propose}. This removes the requirement of using a re-ranker, thereby, saving significantly on time and memory.

{ 
\small
\begin{table}[ht]
\caption{\label{tab2}$N$, $L$, and $V$ for the benchmark datasets.}
\small
\begin{center}
\setlength{\tabcolsep}{3pt}
\begin{tabular}{|l|c|c|c|}
\cline{1-4}
\textbf{Dataset} &${N}$ &${L}$ &${V}$\\
\cline{1-4}
EURLex-4K &15,539 &3,993 &5,000\\
AmazonCat-13K &1,186,239 &13,330 &203,882\\
Wiki10-31K &14,146 &30,938 &101,938\\
WikiSeeAlsoTitles-350K &629,418 &352,072 &91,414\\
AmazonTitles-670K &485,176 &670,091 &66,666\\
\cline{1-4}
\end{tabular}

\end{center}
\end{table}
}

\normalsize

\section{Experiments and Discussion}

\subsection{Experimental Setup}
Code: \url{https://github.com/misterpawan/LightDXML}. 
The experiments are run on a Linux machine with 20 physical cores of {\tt Intel(R) Xeon(R) CPU E5-2640} v4 @ 2.40GHz and 120GB RAM and 2 {\tt RTX 2080Ti} GPU cores. The metrics used for comparison are P@k and PSP@k for $k$=1, 3, 5 \cite{Bhatia16}, train time, average prediction time, and model size. The methods belonging to categories other than the deep learning based methods are all implemented in either {\tt C, C++} or {\tt MATLAB} and are supplied by their respective authors. In particular, AnnexML \cite{annexml} is implemented in C, PfastreXML \cite{pfastrexml} and fastXML \cite{fastxml} in C++, SLEEC \cite{sleec_bhatia} in MATLAB, and Astec and {\algname} in Python. We implement {\algname} in {\tt Python 3.9.4} using {\tt Pytorch 1.8.1} with {\tt CUDA 10.2} and {\tt cuDNN 7.6.5}.

We performed our experiments on several publicly available multi-label datasets in Table \ref{tab2}. We compare {\algname} and other relevant methods belonging to different classes: 1) Embedding based methods, 2) Tree based methods, 3) One-vs-all methods and 4) Deep learning based methods. 
Some Methods such as DECAF \cite{decaf}, ECLARE \cite{eclare} and SiameseXML \cite{dahiya2021siamesexml} are targeted towards exploiting label features, which may not be available in many cases of the XML applications. Thus, a comparison on the LF datasets has not been made against these methods.

Moreover, it is beyond the scope of this paper to compare against all the existing extreme multi-label methods, however, we do highlight results that are best among all those listed in extreme classification repository \cite{Bhatia16}. 

\subsection{Experimental Results} 
\subsubsection{Comparison of methods}
Table \ref{tab_results} lists the precision scores (P@k), propensity based precision scores (PSP@k), training time, prediction time, and total memory for state-of-the-art methods on five datasets in Table \ref{tab2}. As we observe from Table \ref{tab2}, the last two datasets have large number of labels. Except for dataset EURLex-4K, our method has least prediction time on all datasets. In particular, compared to the very recently proposed Astec, our prediction time is two to four times faster. Our model size remains the smallest on three datasets, and second best on two datasets. In particular, compared to Astec, our model requires significantly less memory, yet retaining better accuracy than Astec. A particular case is for dataset AmazonCat-13K, where we require a factor of seven less memory compared to Astec. Although, model sizes and train time of tree-based methods such as Bonsai, Parabel is the least on some datasets, they are known to suffer from poor accuracy as also indicated by lower P and PSP scores in Table. Compared to Astec, we have best P scores on most datasets, except for AmazonTitles-670K, however, we remain very close to Astec's score. Both Astec and \algname are deep learning based methods, and they have highest P scores. When comparing PSP scores, which indicates tail label prediction accuracy, we have best scores on all datasets except for AmazonCat-13K. We would like to point out that recently proposed other methods such as LightXML [10] and APLC-XLNet [23] achieve higher P scores for the first three datasets from top in the Table, but they don’t report train time and their model size is very large, and hence are excluded from comparisons. The boxed and bold values in table indicate best across all methods including those in \cite{Bhatia16}. 

Our method also gives high PSP scores (scores which show tail label relevance) whereas LightXML never shows these scores. For AttentionXML, in larger datasets like AmazonTitles-670K and WikiSeeAlsoTitles-350K, LightDXML outperforms AttentionXML in all aspects, while in EURLex-4K, AmazonCat-13K and Wiki10-31K, LightDXML outperforms AttentionXML in PSP@k while only being slightly behind in P@k. 
Some of the other methods such as LightXML \cite{jiang2021lightxml}, APLC-XLNet \cite{ye2020pretrained} have high model sizes and moreover they do not report PSP scores. Methods like DECAF, ECLARE, SiameseXML and GalaXC \cite{saini2021galaxc} are tailored to use label features on specific datasets like LF-AmazonTitles-1.3M, LF-WikiTitles-500K etc. and hence cannot be compared with our method.

Ablation study is shown in following section. To understand the significant time decrease compared to baseline Astec, we do a detailed module wise analysis in the following subsection.

{
\small 
\begin{table}[h!]
\caption{\label{time_comparison}Timings for each individual module in ASTEC versus LightDXML (in seconds). For LightDXML, Module-1 corresponds to Step-1 and Step-2, Module-2 corresponds to Step-3, and Module-3 and 4 correspond to Step-4 of Algorithm~\ref{alg:propose}. Here M1, M2, M3, M4 stand for modules 1, 2, 3, and 4. Here WT-320K stands for WikiSeeAlsoTitle-320K.}
\bgroup
\setlength{\tabcolsep}{.2em}
\begin{center}
\small
\begin{tabular}{|c|c|c|c|c|c|}
\hline
Dataset & Method &M1 & M2 & M3,4 & Re-ranker\\
\hline
\multirow{2}{*}{EURLex-4K} &Astec  & 67.6 & 4.1 & 43.0 & 24.0 \\ &LightDXML & 21.6 & 7.7 & 45.0 & 0.0 \\
\hline
\multirow{2}{*}{Wiki10-31K} &Astec  & 186.0 & 10.9 & 60.0 & 35.0 \\ &LightDXML & 84.1 & 10.5 & 72.0 & 0.0 \\
\hline
\multirow{2}{*}{WT-320K} & Astec  & 2644.0 & 321.0 & 783.0 & 1321.0 \\ &LightDXML & 303.0 & 136.0 & 963.0 & 0.0 \\
\hline
\end{tabular}
\end{center}
\egroup
\end{table}
}
\normalsize

\subsubsection{Module-wise time comparison and improvement compared to Astec}
% \textcolor{blue}{
We believe that faster prediction time is very relevant in practical applications like online advertisement and hence the community is looking to improve on prediction time. Moreover, if faster prediction time comes with relatively lesser training time then it is also desirable.
% }
To this end, in Table \ref{time_comparison}, we compare the module wise timings for Astec and LightDXML. Referring to Algorithm \ref{alg:propose} for \algname, we recall that Step 1 of LightDXML corresponds to Module 1 of Astec, Step 3 corresponds to Module 2, and Step 4 corresponds to Module 3 and 4 combined. The time comparisons can be explained as follows.
\begin{itemize}
    \item \textbf{Module 1 (intermediate representation learning):} Since LightDXML needs to learn the label representations as a priority over feature representations, a surrogate task is not required, which causes the time taken to be considerably lesser as evident from Table \ref{time_comparison}. As seen in Table \ref{time_comparison}, our method is roughly two to eight times faster compared to Astec.
    \item \textbf{Module 2 (negative sampling):} For Astec, the label representations for approximate nearest neighbor (ANN) training are created by taking the centroid over training points, but LightDXML trains these label representations directly. Thus, on larger datasets, as seen from the table, LightDXML has faster ANN train time.
    \item \textbf{Module 3 and Module 4 (final representation and classifier training):} Since LightDXML repeats the training procedure multiple times after re-learning the label representations, these modules take slightly longer to run for LightDXML. However, each repetitions are much faster as shown above, hence, total time is only slightly more than Astec.
    \item \textbf{Re-ranker:} As discussed earlier, LightDXML does not require a re-ranker, hence, saves significantly on time (and on memory, because it has learnable weights), especially on larger datasets.
\end{itemize}

\subsubsection{Ablation Study}
\begin{figure*}[h]
    \centering
    \includegraphics[width=0.7\textwidth]{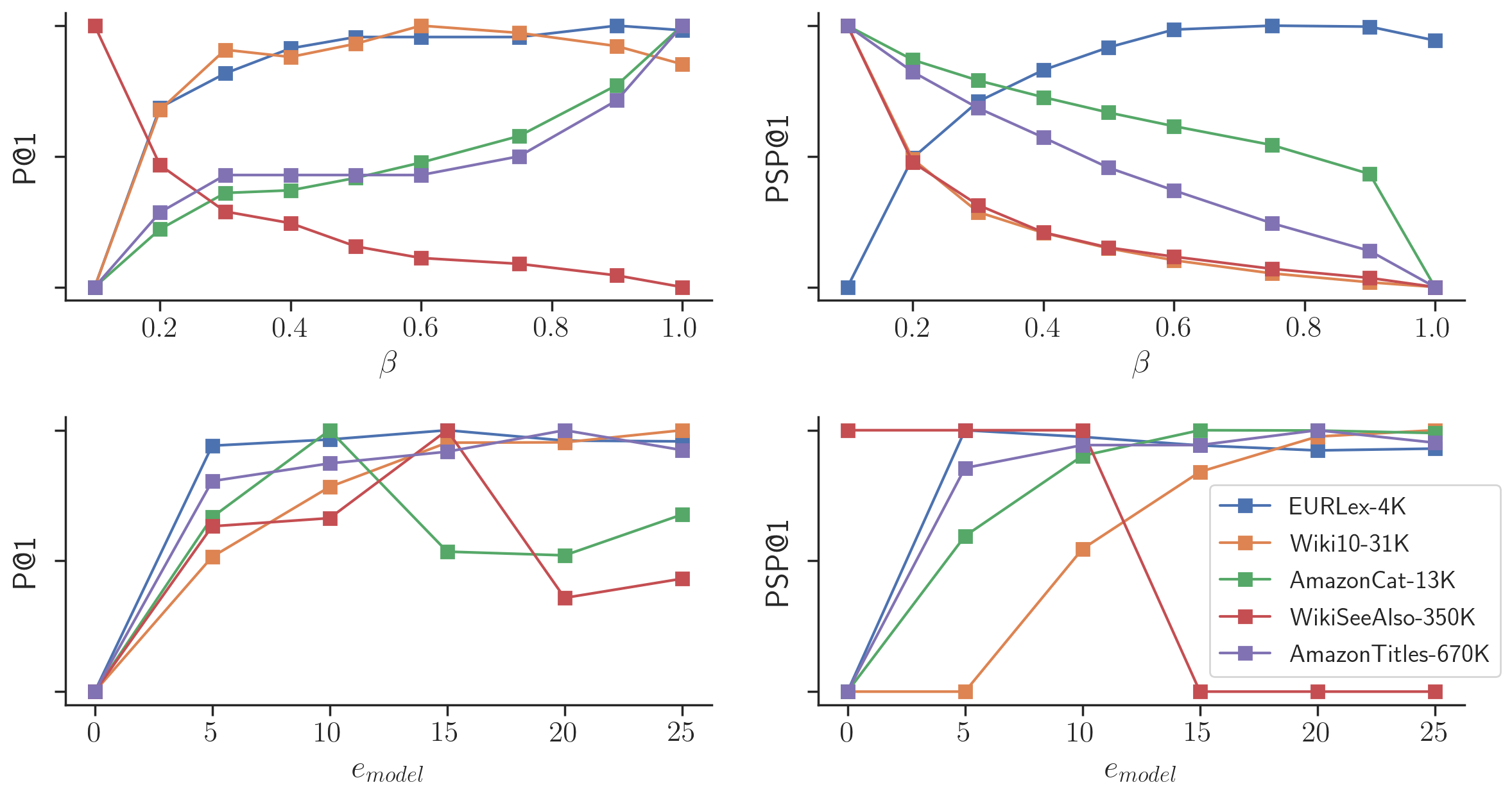}
    \caption{{Plots showing change in P@1 and PSP@1 with respect to the hyperparameters $\beta$ and $e_{\text{model}}$. The plots for each dataset are normalized to fit the different P@1/PSP@1 scores into one plot and only depict the changes.}}
    \label{fig:ds}
\end{figure*}

The hyper-parameters are tuned to minimize the value of the validation loss during training. Ablation study is shown in Figure \ref{fig:ds}. The different hyper-parameters are: $\beta$ as mentioned in Section ``Prediction" in main paper, $k$ is the number of labels considered from the shortlist, 
%the learning rate of the extreme classifier, 
$e_{\text{model}}$ is the number of epochs the extreme classifier trains for, $e_{\text{label}}$ is the number of epochs the label embeddings are trained for each time, and $\hat{e}_{\text{label}}$ is the number of epochs after which the label embeddings are {\it retrained}. 

\small 
\begin{table}[h!]
\caption{\label{tab_hyperparameters}Hyperparameters used for {\algname} on various datasets.}
 \bgroup
 \setlength{\tabcolsep}{1.5pt}
\begin{center}
\begin{tabular}{|c|c|c|c|c|c|c|}
\hline
Dataset &$\beta$ &$k$ &Learn. rate &$e_{\text{model}}$ &$e_{\text{label}}$ &$\hat{e}_{\text{label}}$\\
\hline
EURLex-4K &0.75 &500 &0.006 &16 &8 &8\\
AmazonCat-13K &1 &500 &0.007 &21 &4 &4\\
Wiki10-31K &0.60 &600 &0.009 &21 &8 &6\\
WT-320K &0.1 &500 &0.008 & 21 & 6 & 7\\
% Amazon-670K &0.1 &500 &0.007 &21 &4 &8\\
AmazonTitles-670K &1 &650 &0.007 &21 &6 &9\\
\hline
\end{tabular}
\end{center}
 \egroup
\end{table}
\normalsize

The values of these hyper-parameters that are chosen for the different datasets are listed in Table \ref{tab_hyperparameters}.
It is observed that for smaller datasets, the P@1 scores form a parabolic curve with respect to $\beta,$ but for larger datasets, it is either non-increasing or non-decreasing. Figure \ref{fig:ds} shows the plots of the P@1 scores for various values of $\beta$ ranging from 0 to 1 on various datasets. It was observed in our experiments that the precision scores increase monotonically with increase in the value of $k,$ which is to be expected, but increasing $k$ can lead to increased memory requirements leading to out-of-memory errors, thus, it is kept between 500 and 700. The other hyper-parameters are tuned to minimize the train loss value during training.

Experiments show that the usage of a re-ranker does not provide significant gains in the performance of the proposed model, thereby confirming that the cyclic update procedure used removes the requirement for a re-ranker. On the other hand, using a re-ranker also increases the training and prediction time.

\section{Conclusion and Future Work}
This paper presents a light-weight, simpler, robust, and faster framework {\algname} for extreme classification that has been put to test on several large scale, real world datasets. {\algname} makes use of label embeddings and does not require a re-ranker module. It achieves state-of-the-art precision score and best propensity scores on several datasets, all of this with a factor of $2$ to $3$ speedup in training time (except for tree based methods, which have faster training but poorer P and PSP scores) and prediction time and with a factor of 2 to 3 decrease in memory usage compared to recently proposed state-of-the-art deep learning based method. 

\section*{Acknowledgement}
This work was partially funded by Microsoft MAPG Grant and Seed grant at International Institute of Information Technology at Hyderabad, India. We thank the institute for the HPC facilities that led to completion of this work.

\bibliographystyle{IEEEtran}
\bibliography{LightDXML}

\end{document}